\documentclass[conference,a4paper]{IEEEtran}
\IEEEoverridecommandlockouts

\usepackage[T1]{fontenc}
\usepackage[utf8]{inputenc}
\usepackage{microtype}
\usepackage{graphicx}
\usepackage{booktabs}
\usepackage{tabularx}
\usepackage{array}
\usepackage{enumitem}
\usepackage{xcolor}
\usepackage{xurl}
\usepackage{listings}
\usepackage{hyperref}
\usepackage[nameinlink,noabbrev]{cleveref}

\newcommand{\code}[1]{\nolinkurl{#1}}
\setlist{itemsep=0.25em,topsep=0.4em}

\title{A Governed Agent Architecture for Natural-Language Dashboard Generation over Enterprise DataBrain}

\author{%
\makebox[0.99\textwidth][c]{%
\begin{minipage}[t]{0.24\textwidth}
\centering
{\normalfont\large Yiou Wu\par}
{\normalfont\small
\textit{Department of}\\
\textit{Computer Science}\\
\textit{University College London}\\
London, United Kingdom\\
\href{mailto:yiou.wu.24@ucl.ac.uk}{\nolinkurl{yiou.wu.24@ucl.ac.uk}}\par}
\end{minipage}%
\hfill
\begin{minipage}[t]{0.24\textwidth}
\centering
{\normalfont\large Zezhi Tang\par}
{\normalfont\small
\textit{Department of}\\
\textit{Computer Science}\\
\textit{University College London}\\
London, United Kingdom\\
\href{mailto:zezhi.tang@ucl.ac.uk}{\nolinkurl{zezhi.tang@ucl.ac.uk}}\par}
\end{minipage}%
\hfill
\begin{minipage}[t]{0.24\textwidth}
\centering
{\normalfont\large Ningwei Bai\par}
{\normalfont\small
\textit{Department of}\\
\textit{Computer Science}\\
\textit{University College London}\\
London, United Kingdom\\
\href{mailto:ningwei.bai.23@ucl.ac.uk}{\nolinkurl{ningwei.bai.23@ucl.ac.uk}}\par}
\end{minipage}%
\hfill
\begin{minipage}[t]{0.24\textwidth}
\centering
{\normalfont\large Liuhaichen Yang\par}
{\normalfont\small
\textit{Department of}\\
\textit{Computer Science}\\
\textit{University College London}\\
London, United Kingdom\\
\href{mailto:liuhaichen.yang.23@ucl.ac.uk}{\nolinkurl{liuhaichen.yang.23@ucl.ac.uk}}\par}
\end{minipage}%
}%
\thanks{Zezhi Tang, Ningwei Bai and Liuhaichen Yang are co-corresponding authors.}
}

\hypersetup{
  pdftitle={Text2Dashboard: A Governed Agent Architecture for Natural-Language Dashboard Generation over Enterprise DataBrain},
  pdfauthor={Yiou Wu, Zezhi Tang, Ningwei Bai, Liuhaichen Yang},
  pdfkeywords={large language model agents, natural-language interfaces, dashboards, DataBrain, human oversight},
  hidelinks
}

\begin{document}
\bstctlcite{paper:BSTcontrol}
\maketitle

\begin{abstract}
We present a DataBrain-specific prototype that turns natural-language
analytic requests into inspectable dashboards. An installable Codex plugin and
standalone Agent Runtime combine schema-constrained model decisions with typed
tools, persistent state, and deterministic Hooks for approval, audit,
checkpointing, recovery, and failure handling. The pipeline resolves entities,
discovers metadata, enforces read-only SQL, composes dashboards, and applies
static checks, dynamic preflight, and browser inspection. The model proposes
actions while deterministic software controls execution and records state
transitions.

We evaluate the workflow on frozen real-DataBrain tasks and controlled Hook
faults. Strict success was 6/8 on metadata and SQL tasks: metadata selection
passed 4/4, all four SQL tasks met semantic criteria, and 2/4 met the exact
output-column contract. The final release passed 4/4 single-panel dashboard
tasks, one two-panel task, and one existing-dashboard refinement; a
parameterised task exceeded its step limit. All ten fault scenarios met their
specified outcomes without unapproved external side effects. Model inference
accounted for over 97\% of observed runtime in every reported group. These
small, DataBrain-specific results do not establish production readiness,
general text-to-SQL accuracy,
or an efficiency advantage over manual dashboard construction.

\end{abstract}


\section{Introduction}
\label{chap:introduction}

\subsection{Problem and motivation}

A dashboard request usually starts as a business question rather than a query.
Producing a useful dashboard from that question requires several linked tasks:
identifying the relevant entity and data table, inspecting the schema, writing
SQL, choosing a visual encoding, defining filters, and checking the rendered
result. These tasks are difficult for users who understand the decision they
want to make but do not know the organisation's database structure.

Natural-language generation can reduce this authoring burden, but fluent output
is not sufficient in an enterprise setting. A plausible query may still use a
non-existent field, select the wrong data domain, or request an operation that
the user did not approve. A generated report may also pass a structural check
while failing in the browser. The system therefore needs to control execution
as well as generate content. Its actions must be grounded in current metadata,
restricted to read-only access, recorded for inspection, and recoverable after
an interruption.

Prior work characterises the difficulty of this generation problem. Spider introduced
a cross-domain text-to-SQL benchmark that tests generalisation to unseen
database schemas, while the Big Bench for Large-Scale Database Grounded
Text-to-SQLs (BIRD) increased the scale and realism of
database-grounded tasks~\cite{yu2018spider,li2023bird}. Natural-language
visualisation research adds another stage because a request must be mapped to
an analytic specification and a visual form
~\cite{narechania2021nl4dv,luo2021nvbench,dibia2023lida}. These studies provide
methods and evaluation settings for query and chart generation. They do not
define the permissions, persistent state, approval rules, or recovery behaviour
of this company's dashboard workflow.

\subsection{Project aim and boundary}

The aim of this work is to turn a natural-language analytics request into
an inspectable and dynamically verified dashboard over the company's DataBrain
platform. The system is designed around DataBrain entity identifiers, DataLab
metadata, explicit data domains, BigQuery-oriented SQL, and the existing
dashboard artifact format.

The implemented endpoint is a validated local dashboard preview. Online
DataBrain publication is available in the plugin workflow but is not exposed as
a standalone Runtime tool. A standalone web chat interface is also outside the
evaluated contribution. These boundaries keep the study focused on the Agent
control loop, company data access, and the quality of the generated artifact.

\subsection{System approach}

The system has two entry surfaces. The Codex plugin presents the workflow
through a set of progressively disclosed skills. The standalone Python Agent
Runtime obtains schema-constrained decisions from an authenticated Codex
command-line interface (CLI) session and executes registered typed tools. Both
surfaces reuse the same DataBrain adapters, dashboard state machine, panel
format, HyperText Markup Language (HTML) renderer, and validation pipeline.

The standalone Agent follows a decision, action, and observation loop related
to the Reasoning and Acting (ReAct) interaction pattern~\cite{yao2023react}. In each
iteration, the model may request one registered tool, ask a necessary question,
or finish. The Runtime remains responsible for argument validation, approval,
tool execution, state persistence, and failure handling. This separation is
important because model decisions are probabilistic, while database policy and
artifact state transitions require deterministic enforcement.

The workflow first resolves a game entity and discovers candidate tables. It
then inspects the selected schema before generating bounded, read-only SQL. A
dashboard is built from explicit panel artifacts rather than free-form HTML.
The output passes static validation, an approved dynamic preflight against
DataBrain, and browser inspection. Lifecycle Hooks run at model, tool, checkpoint,
and completion boundaries. They provide exact-call approval, duplicate-call
control, audit logging, checkpointing, dashboard recovery, and bounded handling
of external service failures.

\subsection{Research objectives}

The study has one principal objective: to design and empirically assess a
governed natural-language-to-dashboard agent for the company's DataBrain
environment. Four subsidiary objectives make this aim operational. The first
is to define an architecture that separates probabilistic model decisions from
tool authority, approval, and persistent state. The second is to measure task
correctness for real metadata discovery and approved read-only SQL under
criteria fixed before confirmatory execution. The third is to examine whether
deterministic lifecycle Hooks produce the specified outcomes for approval,
failure handling, checkpointing, audit, and bounded external actions. The
fourth is to assess end-to-end dashboard completion, refinement, execution
cost, and convergence within the implemented prototype boundary.

\subsection{Contributions and evidence}

The central technical contribution is an implemented execution architecture in
which the model proposes schema-constrained actions while deterministic
software retains execution authority through typed tools, lifecycle Hooks, and
durable state. The Codex plugin and standalone Runtime share the
same DataBrain and dashboard capability layer. The system also provides an
inspectable artifact pipeline from verified metadata and read-only SQL to panel
configuration, dynamic HTML, preflight, and local preview. The methodological
contribution is an evaluation procedure that preserves frozen task criteria,
failed runs, staged validation results, timing records, and controlled fault
scenarios.

The evidence characterises a research prototype rather than a production
deployment. Six of eight frozen metadata and SQL tasks met every
strict criterion. All four metadata tasks passed. All four SQL tasks met the
semantic rubric and executed successfully under the read-only policy, while two
failed the exact output-column contract. The final release completed all four
frozen single-panel dashboard tasks. The ten controlled Hook and Runtime fault
scenarios produced their specified outcomes without an unapproved external side
effect. An extension task nevertheless exposed failure to terminate within the
declared step budget. The results are reported by task group because the small,
heterogeneous samples do not justify a pooled reliability estimate.

\section{Background and Related Work}
\label{chap:background}

\subsection{Tool-using language-model agents}

A language-model agent differs from a single text-generation call because it
selects actions, receives observations from an environment, and uses those
observations in later decisions. ReAct demonstrated a prompting pattern that
interleaves model reasoning with actions and observations
~\cite{yao2023react}. Toolformer studied how a language model can learn to
decide when and how to call external APIs~\cite{schick2023toolformer}. These
approaches establish how a model can select and use actions, but they do not by
themselves specify application-level authority over approvals, persistent
artifacts, or recovery. The present system uses the same broad interaction pattern,
but it does not train a new model. A separate Runtime requests one structured
decision at a time and retains authority over the available tools and their
execution.

This distinction changes what should be evaluated. AgentBench evaluates models
in interactive environments where success depends on a sequence of actions
rather than a final answer alone~\cite{liu2024agentbench}. API-Bank tests
planning, retrieval, and calling of external APIs~\cite{li2023apibank}.
ToolEmu studies risks
that arise when language-model agents can use tools with real consequences
~\cite{ruan2024toolemu}. WebArena measures task completion in a reproducible
web environment~\cite{zhou2024webarena}. The $\tau$-bench benchmark evaluates
agents interacting with users and domain-specific API policies, including the
resulting database state~\cite{yao2025taubench}. AgentDojo examines prompt
injection through untrusted tool outputs~\cite{debenedetti2024agentdojo}.
InjecAgent similarly benchmarks indirect prompt injection through external
content encountered by tool-integrated agents~\cite{zhan2024injecagent}.
SkillTV-Bench evaluates whether judges can verify skill-augmented agent
executions using their trajectories and inspectable artifacts
~\cite{han2026skilltvbench}.
These works motivate trajectory-level evidence. Their benchmark and emulation
settings, however, do not constitute evidence that a
particular enterprise workflow enforces its declared control boundaries. The
evaluation therefore records tool order, terminal state,
approval events, latency, and prohibited actions in addition to the generated
SQL and Dashboard artifact.

\subsection{From language to data and visualisations}

Text-to-SQL research addresses the mapping from a question to an executable
query. Spider tests cross-domain generalisation to database schemas that were
not seen during training~\cite{yu2018spider}. BIRD extends this setting with
larger databases, database values, external knowledge, and query efficiency
~\cite{li2023bird}. Decomposed In-Context Learning of Text-to-SQL with
Self-Correction (DIN-SQL) divides the task into schema linking, classification,
query generation, and correction~\cite{pourreza2023dinsql}.
RAT-SQL explicitly models schema relations and links between questions and
database fields~\cite{wang2020ratsql}; a separate schema-linking study finds
that this alignment is a major source of text-to-SQL errors
~\cite{lei2020schemalink}. RESDSQL separates schema linking from SQL skeleton
parsing~\cite{li2023resdsql}. PICARD constrains decoding to reject invalid SQL
prefixes~\cite{scholak2021picard}, while Dr.Spider tests robustness under
changes to questions, schemas, and queries~\cite{chang2023drspider}.
Together, these studies establish schema grounding, constrained generation,
and robustness as components of text-to-SQL evaluation. They do not cover the
exact result-column
contract, approval state, and persistent visual artifact required by the
present workflow.

The present system has a narrower domain and a wider output. It does not attempt to
measure generalisation across unrelated databases. It operates within DataBrain,
where entity identity, data domain, table availability, and permissions must be
resolved before SQL is accepted. The confirmatory benchmark evaluates both
metadata routing and read-only SQL behaviour. It does not support a claim of a
new state-of-the-art text-to-SQL model.

Natural-language visualisation adds analytic and presentation choices.
SeeDB recommends candidate visualisations by scoring patterns in the underlying
data~\cite{vartak2015seedb}.
Data2Vis framed visualisation generation as sequence translation to Vega-Lite
specifications~\cite{dibia2019data2vis}; Vega-Lite defines a grammar for
interactive visual encodings and selections~\cite{satyanarayan2017vegalite}.
Natural Language for Data Visualization
(NL4DV) maps natural-language requests to
attributes, analytic tasks, and candidate charts~\cite{narechania2021nl4dv}.
nvBench provides a cross-domain benchmark for language-to-visualisation tasks
~\cite{luo2021nvbench}. LIDA separates data summarisation, goal exploration, visualisation generation, and
infographic generation
~\cite{dibia2023lida}.
MatPlotAgent combines code generation with visual feedback for scientific
figures~\cite{yang2024matplotagent}.
Data Formulator lets authors specify data concepts for an AI-assisted
visualisation workflow~\cite{wang2024dataformulator}. Data Formulator 2
extends this line of work to iterative authoring with both graphical and
natural-language input~\cite{wang2025dataformulator2}. DynaVis turns natural
language editing requests into persistent visual controls
~\cite{vaithilingam2024dynavis}. VisEval evaluates
generated visualisations beyond code generation alone
~\cite{chen2025viseval}. Together, these works address data representation,
analytic intent, visual specification, and chart evaluation. A valid
specification alone does not show that live company data can be queried safely
or that an interactive report remains valid after parameters and state
transitions are applied.

The present system applies that principle to a company dashboard lifecycle. Each
panel retains its SQL template, parameters, chart configuration, and state.
The final artifact is dynamic HTML rather than a static image. Static contract
validation, DataBrain-backed preflight, and browser inspection remain separate
gates because success at one level does not guarantee success at the next.
\Cref{tab:related-work-position} summarises the relationship between these
research areas and the application-specific requirements of this workflow.

\begin{table*}[!t]
  \centering
  \small
  \caption{Relationship between the closest research areas and the project gap.}
  \label{tab:related-work-position}
  \begin{tabularx}{\textwidth}{>{\raggedright\arraybackslash}p{2.7cm}XX}
    \toprule
    Area & Main focus & Requirement addressed by this work \\
    \midrule
    Tool-using agents & Multi-step decisions and environmental observations
    & Deterministic authority over DataBrain tools, approval, and state \\
    Text-to-SQL & Schema-grounded query generation and execution accuracy
    & Entity and data-domain routing under company permissions \\
    Language-to-visualisation & Analytic specifications and chart generation
    & Persistent panel artifacts and verified dynamic HTML \\
    Agent orchestration & Persistence, interrupts, and human intervention
    & Inspectable lifecycle policies tailored to the DataBrain workflow \\
    \bottomrule
  \end{tabularx}
\end{table*}

\subsection{Human control and lifecycle governance}

Human-AI interaction research emphasises
communicating system capabilities,
supporting correction, and maintaining appropriate user control
~\cite{amershi2019guidelines}. AI Chains shows how decomposing language-model
tasks can expose intermediate outputs and local control points
~\cite{wu2022aichains}. The United States National Institute of Standards and
Technology (NIST) AI Risk Management Framework places validity, safety,
resilience, accountability, transparency, and privacy within a wider
risk-management process~\cite{tabassi2023airmf}.

These sources motivate human control and inspectability at different levels,
but they do not prescribe the execution semantics of the DataBrain workflow.
In particular, a visible confirmation message is weaker than an approval bound
to the exact arguments of one pending external action.

The present system implements a limited engineering interpretation of these ideas.
Approval is bound to the exact tool name and canonical arguments. Checkpoint,
approval, and policy failures stop execution. Audit failure is recorded but is
not allowed to disable the main workflow. These mechanisms address observable
failure modes in this application. They do not demonstrate compliance with a
complete organisational governance framework, and they do not replace the
company's access controls.

\subsection{Position relative to orchestration frameworks and
retrieval-augmented generation (RAG)}

LangGraph documents durable execution, persistence, and human-in-the-loop
interrupts as features of its orchestration runtime
~\cite{langgraph2026overview,langgraph2026interrupts}. The present system implements
overlapping concepts in a small custom Runtime. This choice makes the event
order, state representation, and approval authority directly inspectable for
the dissertation. It is not evidence that a custom Runtime is generally better.
No controlled performance or correctness comparison with LangGraph was
performed.

The standalone Agent does not implement retrieval-augmented generation.
Its runtime dependencies do not include LangChain, LangGraph, an embedding
model, or a vector database. Instead, a deterministic function injects a
compact, source-labelled DataBrain routing summary into the model context,
while physical tables and columns must still be verified through live
metadata tools.

\subsection{Engineering problem addressed}

Prior work provides methods for tool use, schema-grounded SQL, visual
specification, and human intervention. The application-specific engineering
problem addressed here is their controlled integration within DataBrain. The
model must be able to propose metadata and dashboard actions while authority
over permissions, approval, persistence, and validation remains in
deterministic software. The following methodology examines one implementation
of that boundary. It does not claim that the architecture resolves a general
gap across databases or orchestration frameworks.

\section{Methodology}
\label{chap:methodology}
\subsection{Study Design and System Requirements}
\label{chap:requirements}

\subsubsection{Operational context}

The target users are analysts and product managers who can specify an analytic
need without necessarily knowing DataBrain entity identifiers, table names,
SQL, or dashboard configuration. The studied workflow begins with a
natural-language request and proceeds through metadata selection, approval of
real data execution, artifact validation, and local browser inspection. The system
also supports a constrained revision of an existing dashboard while preserving
unrelated panels.

Data owners and administrators impose independent requirements concerning table
permissions and company-data transfer. Runtime recovery and diagnostic traces
are required for implementation review. Experimental assessment further
requires criteria fixed before confirmatory execution, retained failures, and a
separation between implemented mechanisms and observed task outcomes. These
constraints determine the functional and non-functional requirements below.

\subsubsection{System requirements}

\Cref{tab:requirements} summarises the requirements used to assess the
implementation. Functional requirements describe the user-visible workflow.
Non-functional requirements constrain how the workflow is executed. Functional
requirements (FR) and non-functional requirements (NFR) use the corresponding
prefixes in \Cref{tab:requirements}.

\begin{table*}[!t]
  \centering
  \small
  \caption{Evidence-backed system requirements.}
  \label{tab:requirements}
  \begin{tabularx}{\textwidth}{>{\raggedright\arraybackslash}p{1.1cm}X}
    \toprule
    Code & Requirement \\
    \midrule
    FR1 & Accept a natural-language goal and obtain one schema-constrained model decision at a time. \\
    FR2 & Resolve DataBrain entities, discover tables, inspect schemas, and distinguish public from business data scope. \\
    FR3 & Generate bounded read-only SQL, support dry-run, and require exact approval before real execution. \\
    FR4 & Create persistent panel artifacts containing SQL, parameters, chart configuration, and lifecycle state. \\
    FR5 & Generate dynamic HTML and keep static validation, DataBrain preflight, and credentialed preview as separate gates. \\
    FR6 & Persist Agent state, resume eligible runs, and refine an existing Dashboard without unintended replacement. \\
    NFR1 & Reject unknown tools, malformed arguments, SQL mutation, and unapproved external actions. \\
    NFR2 & Retain redacted decisions, tool results, Hook directives, failures, and timing data for audit and recovery. \\
    NFR3 & Keep credentials outside model prompts, persisted state, generated HTML, and dissertation evidence. \\
    NFR4 & Bound retries, table exploration, duplicate calls, and total Agent steps. \\
    NFR5 & Freeze evaluation criteria, retain failed runs, and produce reproducible structured result files. \\
    \bottomrule
  \end{tabularx}
\end{table*}

Approval is grounded in Runtime state rather than model text. It applies to the
fingerprint of one tool name and one canonical argument object. A changed SQL
query, date range, or preflight request therefore requires new approval. A
successful preflight receipt is similarly bound to the hash of one generated
HTML report. Regenerating the report invalidates the receipt.

\subsubsection{Domain boundary}

The DataBrain-specific scope allows the project to define explicit rules for
game identity, public and business data domains, DataLab metadata,
BigQuery-oriented SQL, permission failures, and the company's dashboard format.

The project does not train a language model or claim general text-to-SQL
accuracy across unrelated databases. The standalone evaluated boundary ends at
an approved, dynamically preflighted local preview. Online DataBrain publishing
is described by the plugin workflow but is not a registered standalone tool.
A separate web chat interface, a user study, and a comparison with LangGraph are
also outside the completed evaluation.

\subsubsection{Privacy, ethics, and company data handling}

The model provider delegates authentication to the official Codex CLI using a
ChatGPT account sign-in flow~\cite{openai2026auth}. Text2Dashboard does not read
or offer an OpenAI application programming interface (API) key path. It also removes credential-like environment
variables from the model subprocess. DataBrain credentials remain outside that
model process and are used by the outer tool Runtime and credentialed DataBrain,
preflight, or preview paths.

Real DataBrain observations may enter later model context. Each real run and
cross-process resume therefore requires the explicit
\code{--allow-company-data-to-codex} acknowledgement. The flag records a user
decision but does not grant company permission. Raw runs, SQL results, audit
files, and screenshots remain in ignored local evidence directories. Only
manually reviewed and redacted derivatives may enter the dissertation.

The evaluation did not recruit participants or collect new personal data.
Consequently, no user-study or usability claim is made. Raw SQL results, audit
traces, internal identifiers, credentials, and unredacted screenshots are not
included in the submitted report. The numerical analysis uses locally retained
structured summaries that do not expose row-level company data. This procedure
reduces disclosure in the dissertation but does not replace the company's
access-control and information-governance processes.

\subsubsection{Evaluation design}

The study follows an iterative systems-engineering method. Requirements were
implemented as typed tool schemas, state transitions, Hook policies, and
artifact contracts. The reported evaluation concerns the final frozen release.
Task criteria were fixed before confirmatory execution, and failed outcomes
were retained rather than removed after inspection.

\begin{figure*}[!t]
  \centering
  \includegraphics[width=\textwidth]{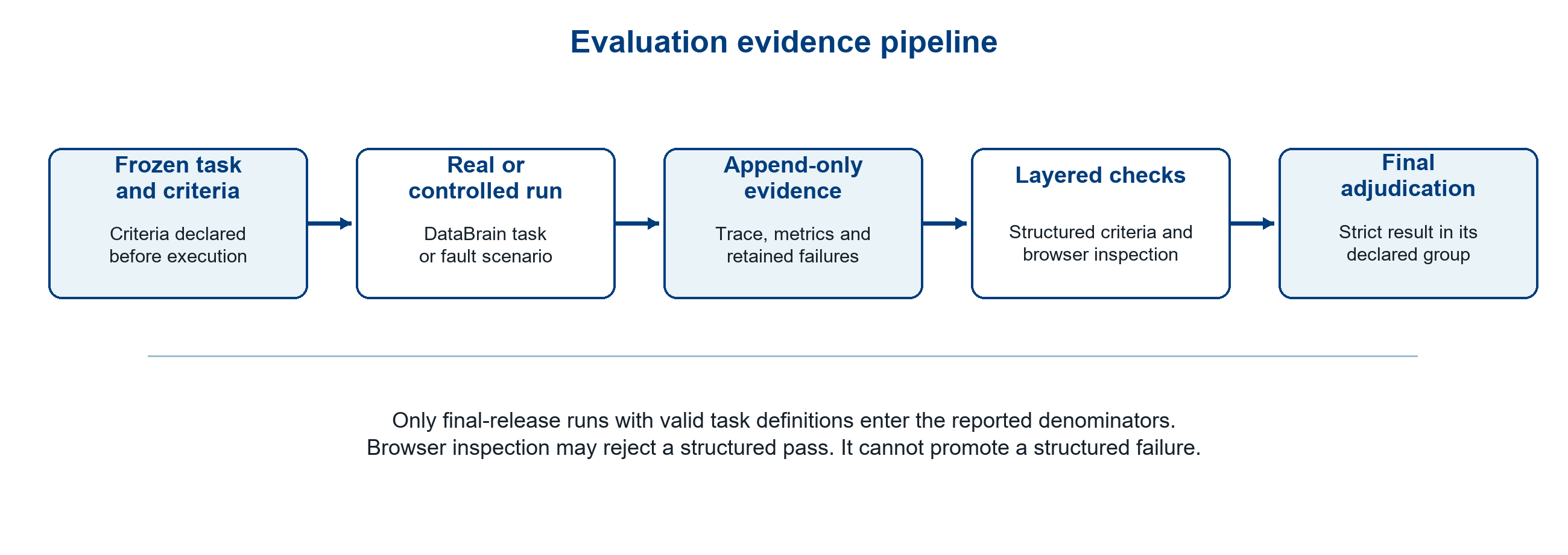}
  \caption{Evaluation evidence pipeline. Frozen criteria and append-only run
  records precede structured checks, rubric-based browser inspection where
  required, and
  final adjudication against a declared denominator.}
  \label{fig:evaluation-pipeline}
\end{figure*}

The evaluation comprises four evidence groups. First, the real-DataBrain core
benchmark contains eight valid tasks, split evenly between metadata and
read-only SQL. Second, four frozen dashboard tasks, labelled D01 to D04,
evaluate key performance indicator (KPI), line, category, and stacked visualisations against structured artifact criteria
and a predefined browser-inspection rubric. Third, the extension evidence
contains one
two-panel task, one joint parameter task, and one existing-dashboard refinement.
Fourth, ten controlled fault scenarios examine permission
errors, gateway retries, schema mismatch, approval, duplicate calls, search
limits, checkpoint failure, changed arguments, and fail-open audit behaviour.
The resulting evidence sequence is summarised in
\Cref{fig:evaluation-pipeline}.

Only tasks with frozen criteria and valid execution records enter a reported
denominator. Infrastructure failures and benchmark-construction errors remain
in the local evidence record but are not analysed as Agent outcomes.

\subsubsection{Outcome measures and analysis}

The primary outcome is strict task success against the frozen task rubric. For
a complete dashboard, success requires correct entity and data scope,
acceptable table and fields, semantically valid read-only SQL, a valid panel,
successful static validation and preflight, a usable browser render, and no
prohibited or unapproved action. Stage-level results are reported separately so
that a correct SQL query cannot hide a failed terminal state or an unusable
dashboard.

Each criterion was evaluated from a specified evidence source. Structured run
records established tool order, approvals, terminal state, timing, and
prohibited actions. Persisted panel and report artifacts established field,
parameter, and lifecycle contracts. Dynamic preflight established compatibility
with live DataBrain responses. Browser inspection assessed visible content and
console state using the predefined rubric. These evidence types were kept
separate so that a favourable observation at one stage could not override a
failure at another.

Efficiency measures include wall time, model time, tool time, model decisions,
tool calls, token counts, and approval interventions. Reliability measures
include error classification, bounded retries, loop prevention, recovery state,
and prohibited side effects. Success proportions use Wilson 95\% confidence
intervals because the task samples are small~\cite{wilson1927probable}. Raw
denominators and individual failures remain visible. No significance test is
reported, and no manual-workflow speed comparison or user study was completed.

These requirements and evaluation rules define what the implementation must
enforce and what the reported observations may support. The following system
design maps them to explicit decision, governance, state, and artifact
boundaries.

\subsection{System Architecture and Design}
\label{chap:architecture}

\subsubsection{Design principles}

Text2Dashboard is organised around a simple division of authority. The
language model proposes the next action, while deterministic code decides
whether that action may run and how its result is recorded. This prevents a
fluent model response from becoming equivalent to approval, valid SQL, or a
verified Dashboard.

Four principles shaped the architecture. First, probabilistic reasoning and
deterministic enforcement are separate. Second, the Codex plugin and the
standalone Agent share domain capabilities rather than maintaining two
implementations. Third, recovery depends on files that survive both model and
Python process failure. Fourth, report generation is staged so that static
correctness, live-data compatibility, and browser inspection remain distinct claims.

\subsubsection{Dual-entry architecture}

\begin{figure*}[!t]
  \centering
  \includegraphics[width=\textwidth]{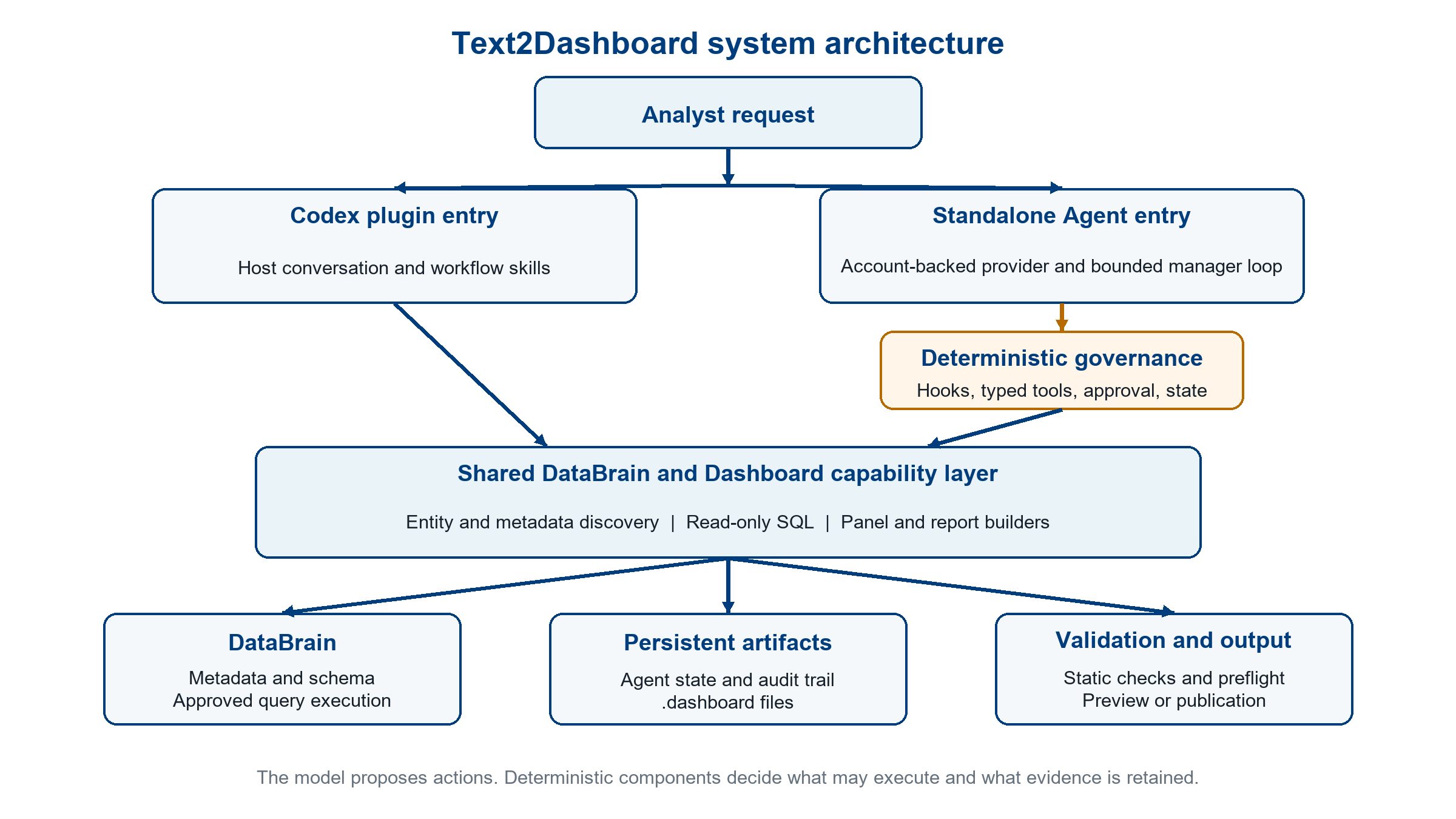}
  \caption{Dual-entry architecture with deterministic authority retained by
  the Runtime, Hooks, tools, and artifact validators.}
  \label{fig:architecture}
\end{figure*}

\Cref{fig:architecture} summarises the two entry points and their shared
DataBrain and Dashboard capability layer.

The plugin is the embedded entry point. Codex provides the surrounding
conversation and invokes skills that impose the Text2Dashboard workflow and
load domain guidance when needed. The standalone package supplies its own
manager loop, typed tools, state, Hooks, command-line resume, and experiment
runner. Both entries reuse the same DataBrain scripts, Dashboard state utility,
composer, validator, and report builder. They are therefore two orchestration
surfaces over one capability layer, not two separate Dashboard products.

The evaluated standalone boundary ends at an approved local preview. Online
DataBrain publication belongs to the plugin publishing workflow and is not
registered as a standalone Agent tool. The shared Dashboard state still
recognises \code{published} so that artifacts remain compatible when control
returns to the plugin.

\subsubsection{Structured decision loop}

The provider interface contains one operation,
\code{decide(state, tool_schemas)}. Its result is an \code{AgentDecision} with
exactly one of three meanings: call one tool, ask one question, or finish. This
narrow protocol keeps ChatGPT authentication and Codex process management
outside the core Runtime and makes each proposed transition inspectable.

For each bounded iteration, \code{AgentRuntime} emits \code{before_model},
requests one schema-constrained decision, records the decision, and emits
\code{after_model}. A tool decision then passes through \code{before_tool}, the
registry, and either \code{after_tool} or \code{tool_error}.
\code{ToolRegistry} validates required fields, rejects additional or mistyped
arguments, and converts exceptions into \code{ToolResult} observations. The
next model turn therefore receives a structured outcome instead of gaining
direct Python, shell, file, or DataBrain access. A maximum-step limit provides
a final bound if other stopping rules do not terminate the run.

The Codex subscription adapter reuses one Codex session within a run. Later
calls send only new Runtime events while the session retains the earlier
instructions and schemas. Session reuse reduces repeated context transfer, but
the session is a performance aid rather than the source of truth. The Runtime
remains authoritative for the trace, allowed tools, approvals, stopping, and
persisted state.

\subsubsection{Hook governance layer}

Hooks place policy at stable lifecycle boundaries instead of relying on the
manager prompt. The normal event order is:

\begin{lstlisting}
run_start -> checkpoint -> before_model -> after_model
          -> before_tool -> after_tool | tool_error -> checkpoint
          -> ... -> run_complete | run_failed -> checkpoint
\end{lstlisting}

Each Hook can return \code{continue}, \code{pause}, \code{block_external}, or
\code{block}. A pause creates a resumable user question. An external block
records a temporary dependency failure separately from an internal Agent
failure. Hook implementations also choose a failure policy. Audit is fail-open
because losing one telemetry write should not silently become a business
decision. Approval, Dashboard-state loading, circuit breaking, loop protection,
and checkpoint persistence are fail-closed because their failure removes a
safety or recovery guarantee.
The responsibilities and execution boundaries of the built-in Hooks are
summarised in \Cref{tab:hooks}.

\begin{table*}[!t]
  \centering
  \small
  \caption{Responsibilities of the built-in governance Hooks.}
  \label{tab:hooks}
  \begin{tabularx}{\textwidth}{>{\raggedright\arraybackslash}p{3.7cm} >{\raggedright\arraybackslash}p{2.0cm} X}
    \toprule
    Hook & Boundary & Responsibility \\
    \midrule
    \mbox{\texttt{DashboardResumeHook}} & Run start & Validate and load existing Dashboard state. \\
    \mbox{\texttt{ToolApprovalHook}} & Tool boundary & Pause before real SQL, preflight, or preview, then consume the approval. \\
    \mbox{\texttt{ToolLoopGuardHook}} & Tool boundary & Stop an unchanged duplicate call and bound table searches. \\
    \texttt{DataBrainCircuit}\newline\texttt{BreakerHook} & Tool error & Convert exhausted metadata gateway retries into an external block. \\
    \mbox{\texttt{CheckpointHook}} & Checkpoint & Atomically persist serialised \code{AgentState}. \\
    \mbox{\texttt{AuditHook}} & All events & Append redacted line-delimited JavaScript Object Notation records without controlling execution. \\
    \bottomrule
  \end{tabularx}
\end{table*}

Approval is bound to a Secure Hash Algorithm 256-bit (SHA-256) fingerprint of the tool name and canonical
arguments. Fields that could imitate confirmation are excluded from the
canonical object. The Runtime adds the internal confirmation flag only after
matching an approval already held in \code{AgentState}. Changing SQL, a date
range, or another argument changes the fingerprint and requires a new approval.
The approval is removed after success or failure, so it cannot authorise a later
call.

The loop guard combines a call fingerprint with a hash of \code{.dashboard}.
This distinction matters during iterative construction. Repeating the same
validator after changing a panel is legitimate, while repeating the same
metadata call with no new evidence is not. The circuit breaker is also
deliberately narrow. Metadata operations retry selected timeouts and Hypertext
Transfer Protocol (HTTP) 502,
503, and 504 responses up to three attempts. SQL execution is not automatically
retried because an uncertain execution outcome should not be duplicated.

\subsubsection{Two-layer state and recovery}

Runtime and Dashboard state have different lifetimes. \code{AgentState}
records the request, phase, iteration, trace, open question, pending and granted
approvals, final response, error, and provider session information held in
\code{context_data}. \code{JsonStateStore} validates run identifiers and writes
through a temporary file followed by atomic replacement. A waiting or
externally blocked run can therefore be loaded by a later command and resumed
with new user input.

The second authority is \code{.dashboard/_state.json}. It records Dashboard
identity, panel ordering and status, and publication binding.
\code{DASHBOARD.md}, \code{ROADMAP.md}, and \code{STATE.md} are readable views
generated by the Dashboard state utility. A panel directory holds
\code{CONTEXT.md}, \code{sql.sql}, and \code{panel.json}. Keeping these
artifacts outside \code{AgentState} allows a new run, the plugin, or a human
reviewer to continue the same Dashboard without replaying an old conversation.

On run start, \code{DashboardResumeHook} copies validated Dashboard state into
the new Agent context. Corrupt state fails closed. Cross-process resume rebuilds
the provider and tool registry, and real company-data runs require
acknowledgement again. The special failed-run retry path accepts only provider
failures. It removes the uncertain Codex session before trying the model turn
again because the provider may have completed a decision that the outer Runtime
did not receive.

\subsubsection{Data and artifact pipelines}

The DataBrain boundary exposes five typed tools.
\code{inspect_databrain_metadata} performs the common read-only sequence of
entity resolution, up to three keyword searches, candidate selection, and
schema retrieval. The four fine-grained tools retain explicit access to entity
search, table discovery, schema retrieval, and SQL execution when targeted
refinement is needed.

Entity identity does not determine data scope. Public intelligence and opinion
requests are routed to \code{game_code=databrain}, even when entity search
returns a game-specific code. A business code is retained only when business
scope is explicit or no public metric rule applies. HTTP 401 is represented as
invalid or expired authentication. HTTP 403 means that the credential may be
valid but lacks access to the selected scope or table. This separation avoids
treating every permission response as a failed login.

The shared artifact lifecycle recognises \code{planned}, \code{drafted},
\code{built}, \code{verified}, and \code{published}. Standalone construction
can move directly from a planned entry to a built panel, while the fuller plugin
discussion workflow may use the drafted stage. \code{build_dashboard_panel}
normalises the panel contract, checks panel identity, chart axes, KPI fields,
parameters, and read-only SQL, then writes the panel artifacts.
\code{validate_dashboard_panel} changes the panel to verified only when
validator output is accepted.

Strict report generation collects verified panels in order, merges parameters,
creates \code{dashboard.json}, and renders SQL templates into
\code{dashboard.html}. Static report validation checks the local contract
without contacting DataBrain. Approved dynamic preflight then executes bounded
option and chart queries against real DataBrain, including default, single,
multiple, and empty selection cases where applicable. It checks returned fields
and unresolved placeholders.

A successful preflight receipt stores the DataBrain host, completion time, and
SHA-256 of the exact HTML report. It stores no credential. Regenerating the
report deletes the receipt, and preview refuses a report whose hash no longer
matches. The managed preview proxy binds to localhost, holds a random shutdown
token, and exposes only non-secret status through the Agent tool.

\subsubsection{Security trade-offs}

The Codex subprocess receives a sanitised environment that removes API billing
variables and credential-like names. It runs with a read-only sandbox and a
strict decision schema. A token read from the recommended project \code{.env}
is explicitly passed only to DataBrain, preflight, or preview paths. An
operating-system token can also be inherited by general Dashboard child
processes, although the compose and static validation scripts do not use it.
Panel identifiers and Dashboard paths are constrained to the configured
workspace.

These measures provide defence in depth, not a formal security proof. SQL
validation is lexical and accepts only one statement beginning with
\code{SELECT} or \code{WITH} while rejecting known mutation keywords. It is not
a full SQL parser. Redaction depends on recognised secret patterns, and the
localhost proxy remains a credential-bearing process while active. These limits
define the evaluated prototype boundary and motivate further security assessment
before wider deployment.

The architecture defines where authority resides. The implementation described
next shows how those boundaries are realised in the Runtime, provider adapter,
Hooks, DataBrain tools, artifact lifecycle, and evaluation records.

\subsection{Implementation}
\label{chap:implementation}

The standalone system is implemented as a Python package under
\code{src/text2dashboard_agent}. The package translates the architectural
boundaries in \Cref{chap:architecture} into a bounded Runtime, a model-provider
adapter, deterministic governance Hooks, DataBrain and Dashboard tools, and
evaluation instrumentation. The main modules are summarised in
\Cref{tab:implementation-modules}. The implementation uses Python data classes,
protocols, subprocess management, JavaScript Object Notation (JSON)
persistence, and a typed-tool layer. It
does not depend on LangChain, LangGraph, or an OpenAI software development kit.

\begin{table*}[!t]
  \centering
  \small
  \caption{Principal standalone implementation modules.}
  \label{tab:implementation-modules}
  \begin{tabularx}{\textwidth}{>{\raggedright\arraybackslash}p{4.0cm}X}
    \toprule
    Module & Main responsibility \\
    \midrule
    \code{models.py} & Serialisable decisions, calls, results, phases, trace events, and run state. \\
    \code{runtime.py} & Bounded model, tool, and observation loop with explicit run transitions. \\
    \code{hooks/} & Approval, circuit breaking, loop protection, audit, and checkpoint policies. \\
    \code{tools/} & Typed adapters for DataBrain access and Dashboard artifact operations. \\
    \code{llm/codex_subscription.py} & Account-backed model process, session reuse, and decision validation. \\
    \code{cli.py}, \code{evaluation.py} & User interaction, recovery, experiment execution, measurement, and export. \\
    \bottomrule
  \end{tabularx}
\end{table*}

\subsubsection{Runtime implementation}

The Runtime implements one model decision per iteration. A decision has exactly
one of three forms: a request to call one registered tool, a necessary question
for the user, or a final response. Tool requests are represented separately
from execution results, and every transition appends a structured trace event.
This representation allows a proposed action to be inspected before the
corresponding handler is reached and avoids reconstructing experimental
trajectories from terminal text.

Tools are exposed through a registry with JSON-compatible argument schemas.
The registry rejects unknown names, missing fields, unexpected fields, and
basic type mismatches before invoking a handler. Domain-specific checks remain
inside the handlers because the local schema validator implements only the
subset needed by the registered tools. A maximum-step limit bounds the complete
loop. Questions, approval pauses, external dependency blocks, internal
failures, and successful completion are represented as distinct persisted
phases rather than as free-form model messages.

The resulting implementation preserves the principal authority boundary. The
model chooses among advertised actions, but it cannot register a new tool,
bypass argument checks, add approval, alter a checkpoint, or assign a terminal
phase directly. These transitions remain functions of Runtime and Hook logic.

\subsubsection{Model-provider integration}

The provider adapter delegates authentication and model execution to the
official Codex command-line client. This supports ChatGPT account sign-in,
non-interactive structured output, a read-only execution sandbox, and session
resume~\cite{openai2026auth,openai2026cli}. Text2Dashboard verifies that the
client is authenticated through the account flow and does not read or copy its
credential files. An API-key billing path is not exposed by the standalone
Agent.

The first model request contains the manager instructions, a redacted Runtime
state, registered tool schemas, and compact DataBrain routing guidance. A
required output schema limits the result to the three decision forms accepted
by the Runtime. Physical tables and fields are not accepted from the guidance
alone and must be established through metadata tools. Invalid JSON, an unknown
decision type, or malformed tool arguments is returned as a provider failure
rather than being treated as an executable action.

Within a run, the provider records the model-session identifier and a trace
cursor. Subsequent decisions resume the session and transmit only Runtime events
added since the previous turn. This reduces repeated context transfer, while
persisted Runtime state remains the source of truth for tools, approvals,
observations, and stopping conditions. Session history is therefore a context
optimisation rather than a recovery or governance mechanism.

Before starting the model subprocess, the adapter removes API-billing variables
and inherited variables whose names indicate tokens, keys, secrets, passwords,
credentials, or authorisation values. This includes DataBrain credentials.
Only non-secret model configuration is stored in the workspace.

\subsubsection{Governance and persistence}

Hooks are registered at stable lifecycle boundaries around model decisions,
tool execution, errors, checkpoints, and completion. Each Hook returns a
directive that either permits progress, pauses for user input, records an
external dependency block, or blocks the run. Hook order is deterministic, and
processing stops at the first directive that prevents continuation.

Failure policy depends on the guarantee provided by the Hook. Audit is
fail-open because a telemetry-write failure should remain visible without
silently changing a permitted benign action. Approval, checkpointing,
Dashboard-state loading, circuit breaking, and loop control are fail-closed
because failure of these mechanisms removes a safety or recovery condition.
Hook errors and directives are written into the same trace used for later
inspection.

Approval is attached to an exact tool call rather than to natural-language
confirmation. The tool name and canonical arguments are serialised and hashed.
Confirmation-like fields are excluded from that canonical object. A matching
grant is consumed after the call, whether execution succeeds or fails. Changing
the SQL, date range, report, or another argument changes the fingerprint and
requires another approval. Duplicate-call control combines the call
fingerprint with Dashboard state, permitting a repeated validation after a
material artifact change while stopping an unchanged loop.

Agent state and Dashboard state are persisted separately. Agent state stores
the request, phase, trace, questions, approvals, errors, and provider-session
metadata. It is written through a temporary file followed by atomic replacement.
Dashboard state stores artifact identity, panel order, panel lifecycle, and
publication binding. This separation permits a later Runtime process, the
plugin, or a reviewer to inspect and continue the artifact without replaying
the original model conversation. Corrupt Dashboard state blocks resume rather
than being reconstructed speculatively.

\subsubsection{DataBrain integration}

The DataBrain adapter reuses the existing project capability for entity search,
table discovery, schema retrieval, and SQL execution. The standalone tool layer
adds typed inputs, error classification, retry bounds, data-scope routing, and
approval enforcement around that capability. A composite metadata tool performs
the common read-only sequence of entity resolution, bounded keyword searches,
candidate selection, and schema retrieval. Fine-grained metadata tools remain
available when a targeted correction is required.

Entity identity and data scope are treated as separate decisions. Explicit
public-intelligence or opinion requests use the public DataBrain scope, whereas
explicit business requests retain the appropriate game code. This rule is
enforced by the adapter rather than relying only on the model prompt. Metadata
operations retry selected connection failures, timeouts, and HTTP 502, 503, and
504 responses within a fixed bound. HTTP 401 is classified as an authentication
failure, while HTTP 403 is classified as insufficient permission for the
selected scope or table. Real SQL execution is not retried automatically.

The SQL policy removes comments, permits one statement beginning with
\code{SELECT} or \code{WITH}, masks string literals for inspection, and rejects
recognised mutation or administrative keywords. The same policy is applied to
direct execution and to SQL templates stored in panels. A real, non-dry-run
request additionally requires the Runtime-supplied approval marker. This
lexical policy provides a deterministic prototype boundary but is not
equivalent to parser-based SQL verification.

\subsubsection{Dashboard artifact lifecycle}

Dashboard construction is based on persistent panel artifacts rather than
model-generated HTML. A panel contains source context, a SQL template,
parameters, chart configuration, and lifecycle state. Panel identifiers and
workspace paths are constrained before files are written. The builder
normalises supported input variants, checks chart and parameter contracts, and
applies the read-only SQL policy to main, tab, and option queries. A panel is
marked as verified only after its artifact validator succeeds.

Report composition collects verified panels in their recorded order, merges
parameters by name, and renders the shared dynamic HTML template. Static
validation checks chart contracts, parameter declarations, SQL placeholders,
API construction, and credential handling without contacting DataBrain. This
stage is intentionally distinct from dynamic preflight because local structural
validity does not establish compatibility with live results.

Approved preflight executes bounded parameter-option and chart queries against
DataBrain. Where applicable, it covers default, single, multiple, and empty
selections and checks returned fields and unresolved placeholders. Success
creates a receipt bound to the hash of the exact HTML report and containing no
credential. Regenerating the report invalidates the receipt. Local preview is
permitted only when the current report hash matches a successful preflight and
is served through a localhost process with workspace-local lifecycle state.
These rules implement the transition from built artifact to verified dynamic
preview without allowing model text to assert that a stage has passed.

\subsubsection{Evaluation instrumentation}

The experiment runner loads frozen task definitions, rejects inconsistent task
policies, and creates an isolated workspace and persisted state for each run. A
manifest records task order, configuration, random seed, and the applicable
policy. A dedicated Hook blocks any tool declared forbidden for a task before
its handler can execute.

Outcome derivation uses terminal phase, selected metadata, required tool order,
SQL fragments, panel count, artifact contracts, and prohibited attempts. Timing
records distinguish model duration, tool duration, and wall time. Model
attempts, tool calls, token usage, session reuse, approvals, and Hook
interventions are also retained. Raw run results and browser-inspection records
are append-only, while tabular and summary outputs are regenerated from those
records. Browser inspection can reject a structurally successful Dashboard but
cannot promote a run that failed its structured criteria.

\subsubsection{Implementation limitations}

The local argument validator implements a shallow schema subset, and SQL policy
is lexical rather than parser-based. Tool execution is synchronous, secret
redaction is pattern-based, and managed preview depends on explicit subprocess
lifecycle control. A DataBrain token exported as an operating-system variable
may be inherited by operational Dashboard subprocesses, although the model
subprocess remains sanitised. The standalone Runtime does not expose online
publication or a browser chat interface. These are implementation limits of the
evaluated prototype rather than completed or evaluated capabilities.

\section{Results and Analysis}
\label{chap:results}
\subsection{Evidence and adjudication}

The final evaluation uses the four evidence groups defined in
\Cref{chap:requirements}. The real core benchmark evaluates metadata and
read-only SQL. The final single-panel group evaluates four dashboard
archetypes. Extension and refinement tasks examine multi-panel construction,
parameters, termination, and preservation of an existing Dashboard. A
controlled fault experiment evaluates governance outcomes.

Every reported result is assigned to its own declared denominator. Only final
release runs with valid task definitions and experiment infrastructure enter
the analysis. The predefined browser rubric can reject a run that passes
the structured artifact criteria, but it cannot promote a run that fails those
criteria. Wilson 95\% confidence intervals are reported for small binomial task
groups~\cite{wilson1927probable}. No significance test, ablation claim, or
manual-workflow speed comparison is made.
\Cref{tab:evaluation-outcomes} reports the resulting task-group outcomes.

\begin{table*}[!t]
  \centering
  \small
  \caption{Strict task outcomes. Groups are reported separately.}
  \label{tab:evaluation-outcomes}
  \begin{tabularx}{\textwidth}{>{\raggedright\arraybackslash}p{3.0cm} >{\raggedright\arraybackslash}p{2.95cm} >{\centering\arraybackslash}p{1.7cm} >{\centering\arraybackslash}p{1.65cm} X}
    \toprule
    Evidence group & Role & Completed & Strict success & Wilson 95\% interval \\
    \midrule
    Core metadata and SQL & Four metadata and four SQL tasks & 8/8 & 6/8, 75.0\% & 40.93\% to 92.85\% \\
    Final single-panel Dashboard & KPI, line, bar, and stack & 4/4 & 4/4, 100.0\% & 51.01\% to 100.0\% \\
    Dashboard extension & Two-panel task (D06) and joint parameterised task (D05/D07) & 1/2 & 1/2, 50.0\% & 9.45\% to 90.55\% \\
    Existing-Dashboard refinement & Refinement task (R03) & 1/1 & 1/1 observed & Single observation \\
    \bottomrule
  \end{tabularx}
\end{table*}

\begin{figure*}[!t]
  \centering
  \includegraphics[width=\textwidth]{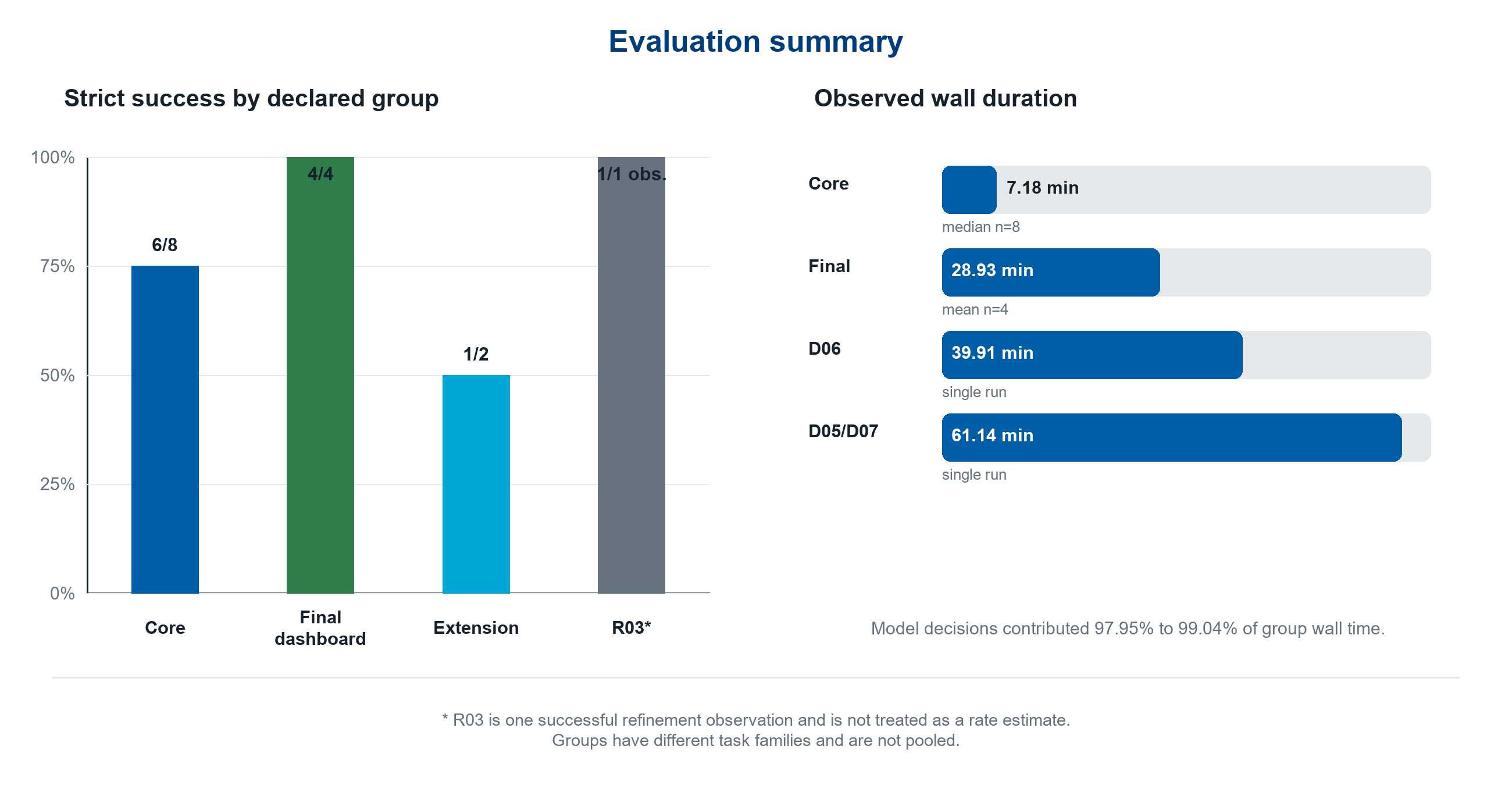}
  \caption{Strict success by declared task group and observed execution time.
  Bars use the denominators shown on the figure. Core time is the median over
  eight runs, the final single-panel group uses a group mean, and D06 and
  D05/D07 are single-run observations.}
  \label{fig:evaluation-summary}
\end{figure*}

The groups are not pooled because they cover different task families and step
budgets. The reported proportions therefore describe only their declared task
sets. \Cref{fig:evaluation-summary} presents the success proportions and
observed execution times without combining the groups.

\subsection{Metadata and read-only SQL}

The valid core denominator contains four metadata and four SQL tasks. Metadata
strict success was 4/4. All selected top-1 entity identifier, entity type, and game-code
checks passed. The six tasks that required table and schema evidence also found
the frozen table and field criteria. No prohibited tool was attempted.

Strict SQL success was 2/4. However, all four SQL tasks passed the semantic
rubric and eventually executed an approved read-only query successfully. The
two strict failures were output-contract failures. One returned
\code{revenue_usd} rather than the frozen alias \code{revenue}. The other
returned \code{total_mentions} rather than \code{mentions}. That task also
produced an invalid fully qualified identifier on its first attempt, observed
the DataBrain error, changed the SQL, obtained a new fingerprinted approval,
and succeeded on the repaired query.

The disaggregated results distinguish semantic query correctness from interface
compatibility. Metadata routing passed in this four-task subset and the SQL
semantic rubric passed in 4/4 cases. Exact result-interface compliance was 2/4,
yielding overall strict success of 6/8. The alias failures affect downstream
Dashboard bindings even when the query is semantically acceptable. They
therefore remain strict failures under the frozen rubric.

\subsection{Final single-panel Dashboard correctness}

The final release was evaluated on four frozen single-panel archetypes: KPI,
line, category bar, and sentiment stack. All four runs passed the declared
entity, metadata, panel, SQL, tool-order, validation, preflight, and
prohibited-action criteria. Each local browser preview was then inspected
against the frozen visual rubric. Every task displayed one visible non-empty
result of the requested form, and no page-console warning or error was observed.
Strict success was 4/4.

Every single-panel task also made one rejected
\code{build_dashboard_panel} call using \code{sql} rather than
\code{sql_template}. The typed tool prevented an invalid
artifact, and the Agent repaired the call on its next decision. This preserved
correctness but added one model round trip per task, exposing a remaining
mismatch between model tool affordance and the formal panel contract.

\subsection{Extension, parameters, and refinement}

D06 required two coherent panels over one data source and time window. It
passed the structured checks, one approved real preflight, and browser
inspection. The output contained exactly two non-empty panels with the required metric and
sentiment-series contracts. This observation establishes successful
multi-panel composition for the evaluated task without manual file editing. It
does not estimate success across other multi-panel requests.

The joint D05/D07 task tested an interactive filter and its default, single,
multiple, and empty selection paths. Its entity, schema, panel, SQL, safety,
and tool-order criteria were observed as correct. Three separately approved
dynamic preflights succeeded. The manager loop then rebuilt the panel after each
valid chain, invalidated the latest report receipt, and did not return a final
decision within 30 steps. The formal outcome therefore remains a failure. This
distinction is important: the parameter and preflight mechanisms produced the
required intermediate states, but the Agent did not recognise that the task was
complete. The observation identifies convergence and termination policy,
rather than parameter execution, as the failed system property in this run.

R03 started from a copied final-release Dashboard and requested only a shorter
time window. The source artifact remained unchanged. The working copy preserved
panel count, panel ID, chart type, metric, entity filter, grouping, and ordering,
while changing the 30-day and 31-row settings to 14 days and 15 rows. Real
preflight and browser inspection passed, and no initialise, add-panel, preview, or
publication action was attempted. This single observation supports
state-preserving refinement, not general multi-turn reliability.

\subsection{Efficiency}

The observed duration and interaction counts are reported in
\Cref{tab:efficiency}.

\begin{table*}[!t]
  \centering
  \small
  \caption{Observed end-to-end efficiency. Aggregate and single-run rows are
  labelled separately.}
  \label{tab:efficiency}
  \begin{tabularx}{\textwidth}{>{\raggedright\arraybackslash}p{2.1cm} >{\raggedright\arraybackslash}p{2.6cm} >{\centering\arraybackslash}p{1.8cm} >{\centering\arraybackslash}p{1.8cm} >{\centering\arraybackslash}p{1.5cm} X}
    \toprule
    Group & Basis & Wall time & Model share & Model attempts & Tool calls \\
    \midrule
    Core & Median or mean over 8 & 7.18 min median & 97.95\% & 3.625 mean & 2.625 mean \\
    Final single-panel & Mean over 4 & 28.93 min & 99.04\% & 15.0 mean & 14.0 mean \\
    D06 & Single run & 39.91 min & 99.00\% & 20 & 19 \\
    D05/D07 & Single run & 61.14 min & 98.45\% & 30 & 30 \\
    \bottomrule
  \end{tabularx}
\end{table*}

Core median wall time was 430.852 seconds, with an interquartile range of
385.413 seconds and p95 of 700.430 seconds. Metadata median time was 242.255
seconds, compared with 631.793 seconds for SQL. The mean tool time was 9.141
seconds per task, while mean model time was 438.834 seconds.

The same pattern became stronger for Dashboard work. Final single-panel tools
averaged 16.31 seconds in total per run, while model decisions averaged 28.65 minutes. D06
spent 23.45 seconds in tools and 39.52 minutes in model decisions. The failed
D05/D07 run spent 56.32 seconds in tools and 60.19 minutes in model decisions.
R03 used 13 model attempts, 11 tool calls, and about 26.1 minutes of active time
excluding explicit approval wait. Its cumulative input usage reached 2.36
million tokens, of which 1.72 million were cached.

These measurements do not establish that the Agent is faster than a human
because no manual baseline was collected. Within the recorded runs, model time
accounted for the largest share of wall time, while DataBrain and local tool
execution accounted for a small share. The observed efficiency constraints
were repeated sequential model decisions and, in D05/D07, failure to terminate
within the step budget. The latency is therefore primarily an orchestration
bottleneck in the evaluated architecture. Each additional repair, validation,
or completion decision incurs another model round trip even when the associated
local or DataBrain tool completes quickly.

\subsection{Controlled Hook fault evaluation}

Ten controlled component scenarios were executed. All ten produced the
specified governance outcome, and no unapproved external side
effect occurred. \Cref{tab:faults} groups the observed outcomes by governance
function.

\begin{table*}[!t]
  \centering
  \small
  \caption{Grouped outcomes from the controlled Hook fault experiment.}
  \label{tab:faults}
  \begin{tabularx}{\textwidth}{>{\raggedright\arraybackslash}p{3.0cm} >{\centering\arraybackslash}p{1.5cm} X}
    \toprule
    Scenario class & Outcome & Observed result \\
    \midrule
    Access and gateway handling & 2/2 & HTTP 403 was not retried, and exhausted HTTP 504 retries entered \code{blocked_external}. \\
    Missing schema evidence & 1/1 & A completed trace without the required field was strictly rejected. \\
    Exact approval & 3/3 & SQL, preflight, and preview paused, and changed arguments required new approval. \\
    Loop control & 2/2 & An unchanged duplicate and a fourth table search were stopped before execution. \\
    Persistence and audit failure & 2/2 & Checkpoint failure blocked the run, while fail-open audit failure preserved one benign action and recorded \code{hook_error}. \\
    \bottomrule
  \end{tabularx}
\end{table*}

Within the registered tools and injected fault classes, the matrix provides
direct evidence for deterministic approval, bounded metadata failure,
observable fail-open telemetry, and fail-closed recovery controls. It does not
establish resistance to arbitrary prompt injection, security against an
adversarial bypass of the lexical SQL policy, or an end-to-end reliability rate
under live service outages.

\section{Discussion}
\label{chap:discussion}
\subsection{Interpretation of findings}

The implementation satisfies the architectural objective of separating model
proposals from execution authority. Model output is constrained to one typed
decision, while the Runtime and Hooks retain control over tool registration,
argument validation, approval, checkpointing, and terminal state. Agent state
and Dashboard state are persisted separately. Consequently, recovery and
artifact continuation do not rely exclusively on the model provider's
conversation history. The plugin and standalone entry points reuse the same
DataBrain and artifact capabilities, although their orchestration mechanisms
remain distinct.

The core benchmark indicates that metadata selection and semantic SQL were
more reliable than exact result-interface compliance in the evaluated sample.
Strict success was 6/8, comprising 4/4 metadata passes and 2/4 strict SQL
passes. All four SQL tasks met the semantic rubric, but two returned column
aliases that did not satisfy the frozen downstream contract. This difference
is operationally relevant because a semantically valid query may still be
unusable by a panel that expects a specific field name.

The controlled fault matrix provides evidence that the implemented Hook
boundaries produced the specified outcomes for ten declared scenarios. The
observations include exact-call approval, changed-argument reapproval, bounded
metadata retries, loop interruption, checkpoint failure, and fail-open audit
behaviour. Zero unapproved external side effects were recorded in these
controlled scenarios. This finding is limited to registered tools and injected
fault classes. It is not a general security assurance.

End-to-end dashboard evidence was mixed. The final release passed four
frozen single-panel tasks. One two-panel task and one existing-dashboard
refinement also passed their structured and browser criteria. In contrast, the
joint parameter task failed because the manager loop did not terminate within 30
steps, despite producing three successful intermediate preflights. Across the
recorded task groups, model decisions accounted for more than 97\% of wall
time. These observations locate the principal remaining limitations in exact
contract adherence, sequential model interaction, and termination control.

Taken together, the evidence supports the architectural claim that model
proposals can be separated from deterministic execution authority, and it
documents successful completion of the specified tasks. It does not justify a
claim of production readiness, a general reliability rate, or autonomous
suitability for unattended daily reporting.

\subsection{Threats to validity}

Internal validity is limited by model and service variability and by one
confirmatory execution per frozen task. The system was developed and evaluated
by the same researcher, and the browser-based visual assessment was performed
by a single evaluator using a predefined rubric. This design introduces
potential experimenter and evaluator bias. The R03 infrastructure correction
was confined to process search paths and preserved the initial artifact hash,
but it still required an exclusion decision.

Construct validity is affected by exact alias requirements, browser inspection,
and step-budget definitions. Alias failures correctly measure interface
compatibility, but they make semantic SQL success look different from strict
success. The frozen visual rubric provides consistent criteria but does not
measure usability as a user study would. A maximum-step failure operationalises
autonomous convergence, even when an intermediate artifact satisfies the
dynamic checks. Other definitions of completion could produce a different
classification.

External validity is limited by a DataBrain-only system, a small set of
evaluated game entities, public intelligence and opinion tables, and a small
set of chart archetypes. Business-table permissions vary independently from
entity visibility. Results therefore do not transfer automatically to other
DataBrain domains, arbitrary databases, schemas, or business intelligence (BI)
platforms. Prior text-to-SQL evaluation work similarly shows that test-set
construction changes what generalisation claims an experiment can support
~\cite{finegan2018evaluation}.

Conclusion validity is limited by small denominators and one confirmatory
execution per task. The wide Wilson intervals make this uncertainty explicit,
and single observations are reported without treating them as rate estimates.
No hypothesis test is reported, no ablation identifies the causal effect of the
composite tool or session reuse, and no manual baseline supports a comparative
efficiency claim. No user study supports a usability claim. Raw company-data
traces and screenshots remain local and are not used as submitted dissertation
figures. The numerical results derive from structured run summaries and
append-only browser-inspection records.

\section{Future Work}
\label{chap:future-work}
The most immediate engineering priorities are therefore targeted improvements to
output-contract adherence and deterministic convergence. The Runtime should
recognise when the current artifact already satisfies a successful preflight and
prevent unnecessary rebuilds before finalisation. Tool descriptions and argument
contracts should also reduce the repeated \texttt{sql} versus
\texttt{sql\_template} repair. Further implementation work should reduce
sequential model calls, adopt parser-backed SQL validation, minimise credential
inheritance by operational subprocesses, and improve preview-process cleanup.
Because model decisions accounted for more than 97\% of observed runtime in the
reported groups, reducing unnecessary model round trips is likely to be an
important direction for improving practical responsiveness.

Future evaluation should strengthen the evidence base rather than only expand
functionality. Representative tasks should be repeated across multiple runs to
measure variability and convergence more convincingly. A small governance
ablation could compare the full Runtime with reduced variants that remove, for
example, approval or loop-control mechanisms, allowing their contribution to
redundant calls, unapproved or prohibited actions, and termination behaviour to be examined more
directly. Broader evaluation should also include additional entity and schema
families, independent visual reviewers, and a predefined manual baseline where
comparative efficiency is a research question.

A further research direction is to examine how much of the current governed
execution design transfers beyond DataBrain. The present study deliberately
uses application-specific metadata, permissions, artifact formats, and
validation rules, so it does not establish portability to other databases or
business-intelligence platforms. Future work could separate reusable governance
mechanisms from DataBrain-specific adapters and evaluate whether the same
authority, state, and validation principles remain effective in other
enterprise analytics environments. Online publication and a standalone browser
interface should likewise be evaluated only after they are implemented as
explicit, approval-gated capabilities.

\section{Conclusion}
\label{chap:conclusion}
This dissertation presented and evaluated Text2Dashboard, a natural-language
Dashboard Agent in which the language model proposes actions while deterministic
software retains authority over DataBrain access, approved read-only SQL,
persistent visual artifacts, and lifecycle governance. The architecture combines
schema-constrained model decisions with typed tools, exact-call approval,
persistent Agent and Dashboard state, staged validation, and controlled recovery.

Within the evaluated task set, metadata routing passed in 4/4 tasks, the final
single-panel release passed 4/4 tasks, one two-panel task passed, and one
state-preserving refinement passed. The ten controlled Hook and Runtime fault
scenarios produced their specified outcomes without an unapproved external side
effect. At the same time, exact SQL output contracts passed in only 2/4 SQL
tasks, the joint parameterised task did not terminate within its step budget,
and new Dashboard tasks required approximately 29 to 61 minutes. These values
describe the declared tasks and should not be interpreted as general reliability
or efficiency rates.

The evaluation highlights two broader lessons from the prototype. First,
semantic correctness alone is insufficient for an artifact-producing agent:
although all four SQL tasks satisfied the semantic rubric, two failed because
their output aliases did not match the downstream panel contract. Evaluation
of this type of system should therefore consider machine-readable interface
compatibility in addition to query semantics. Second, successful intermediate
tool execution does not guarantee agent-level convergence. In the parameterised
task, valid panels and successful preflights were produced, but the manager loop
continued rebuilding the artifact and eventually reached the step limit.
This suggests that completion and termination should be treated as explicit
system properties rather than left entirely to model judgement.

\IEEEtriggeratref{17} 
\bibliographystyle{IEEEtran}
\bibliography{references}

\end{document}